\documentclass[10 pt, conference]{IEEEtran}

\usepackage{url}

\usepackage{amsmath} % assumes amsmath package installed
\usepackage{amssymb}  % assumes amsmath package installed
\usepackage{bm}
\usepackage{graphicx}

\DeclareMathOperator*{\argmax}{arg\,max}
\newcommand{\vect}[1]{\bm{#1}}
\newcommand{\Tx}[1]{\mathcal{T}_#1(\vect{x})}
\newcommand{\thead}[1]{\textbf{#1}}
\newcommand{\thlt}[1]{\mathit{#1}}
\newcommand{\thltb}[1]{\mathbf{#1}}

\title{
Decision-forest voting scheme for classification of rare classes in network intrusion detection
} % maybe make title about people and security (infected users)

\author{\IEEEauthorblockN{Jan Brabec\IEEEauthorrefmark{1}\IEEEauthorrefmark{2} and
Lukas Machlica\IEEEauthorrefmark{1}}
\IEEEauthorblockA{\IEEEauthorrefmark{1}Cisco Systems, Inc., Charles Square Center, Karlovo Namesti 10 Street, Prague, 12000, Czech Republic
}
\IEEEauthorblockA{\IEEEauthorrefmark{2}Czech Technical University in Prague, Faculty of Electrical Engineering, Czech Republic\\
janbrabe@cisco.com, lumachli@cisco.com}
}

\begin{document}

\maketitle
%\thispagestyle{empty}
%\pagestyle{empty}

%%%%%%%%%%%%%%%%%%%%%%%%%%%%%%%%%%%%%%%%%%%%%%%%%%%%%%%%%%%%%%%%%%%%%%%%%%%%%%%%
\begin{abstract}

In this paper, Bayesian based aggregation of decision trees in an ensemble (decision forest) is investigated. The focus is laid on multi-class classification with number of samples significantly skewed toward one of the classes. The algorithm leverages out-of-bag datasets to estimate prediction errors of individual trees, which are then used in accordance with the Bayes rule to refine the decision of the ensemble. The algorithm takes prevalence of individual classes into account and does not require setting of any additional parameters related to class weights or decision-score thresholds. Evaluation is based on publicly available datasets as well as on an proprietary dataset comprising network traffic telemetry from hundreds of enterprise networks with over a million of users overall. The aim is to increase the detection capabilities of an operating malware detection system. While we were able to keep precision of the system higher than 94\%, that is only 6 out of 100 detections shown to the network administrator are false alarms, we were able to achieve increase of approximately 7\% in the number of detections. The algorithm effectively handles large amounts of data, and can be used in conjunction with most of the state-of-the-art algorithms used to train decision forests. \footnote{\copyright~2018 IEEE. Personal use of this material is permitted. Permission from IEEE must be
obtained for all other uses, in any current or future media, including
reprinting/republishing this material for advertising or promotional purposes, creating new
collective works, for resale or redistribution to servers or lists, or reuse of any copyrighted
component of this work in other works. Published paper at IEEE SMC 2018: DOI 10.1109/SMC.2018.00563~\cite{ieeethis}}

\end{abstract}

%\begin{keywords}
%classification, imbalanced data, decision forests, Bayes' theorem 
%\end{keywords}

%%%%%%%%%%%%%%%%%%%%%%%%%%%%%%%%%%%%%%%%%%%%%%%%%%%%%%%%%%%%%%%%%%%%%%%%%%%%%%%%
\section{INTRODUCTION}

Imbalances in distribution of samples across different classes may lead to major performance issues of a classification system, because the system naturally tends to favor majority classes resulting in minority classes being misclassified more often in favor of the majority classes. The imbalance problem can not be neglected since it is inherent to tasks of significant importance. Noteworthy to mention are the areas of security, malfunction detection or processing of medical data \cite{sun2009imbalance}. Note that most of the practical image recognition tasks are intrinsically imbalanced since the object of interest forms only a minor part of the processed image. Recently, the most dominant and extensive is the problem of intrusion detection in network telemetry. Apart from the huge disproportion of benign and malware-related network traffic, the task is coupled with additional strong requirements that have to be met in order to deploy a reliable Network Intrusion Detection System (NIDS). 

First of all, \textit{high precision} of detection has to be achieved since security incidents, produced by NIDS, are in majority of cases assigned to human network administrators and analysts who have to take proper actions. Example of these actions are: analyze the impact of infection, remove the threat or reimage the device. Each of these actions has a different cost, therefore it is crucial to correctly recognize impact and risk of the threat. While devices infected with shady ad-injectors decrease the user experience and may lead to serious infections in the future, information stealing has direct and serious impact on the business of a company. This leads to a \textit{multi-class} setup in which each threat category is represented as a different class. Since human analysts are involved, \textit{explainability and interpretability} of the detections are essential.

In this paper we address NIDS operating on hundreds of enterprise networks, processing tens of millions of network requests per day. Classifier that meets all the requirements outlined in the previous paragraph with an additional benefit of being robust to missing values, is the random forest classifier \cite{Breiman:2001:RF:570181.570182}.  % are the requirements obvious to the reader?
Random forests are one of the best out-of-the-box machine learning solutions \cite{JMLR:v15:delgado14a} well suited for parallel and distributed processing.

% Mention bagging and out-of-bag dataset?

Random forest classifier is an ensemble of decision trees with randomness involved in the learning phase of the classifier \cite{Breiman:2001:RF:570181.570182}. The randomness is important to ensure that the ensemble of the trees is diverse. Diversity of the ensemble is especially useful in the presence of imbalanced data~\cite{diez2015diversity}. Standard ensemble voting scheme used in the prediction phase, is {\it majority voting} \cite{Krawczyk2016}, where the predictions of individual trees are treated as votes and the prediction of the decision forest is determined by the majority of votes. This hard voting scheme has its soft alternative, in which each tree outputs a probability distribution over all class labels. However, it was empirically shown that there is no significant difference in terms of performance between majority and soft voting \cite{brei94}.
%An alternative to majority voting is {\it soft voting} \cite{Zhou:2012:EMF:2381019}, where instead of a hard decision, each tree outputs a probability distribution over all class labels. The prediction of the forest is obtained by averaging the results of individual trees. It was empirically shown that there is no significant difference in results between majority and soft voting \cite{brei94}.

% better emphasise why  imbalanced data are important

Standard aggregation algorithm does not take the prevalence of specific classes into account. In the task of network-traffic based intrusion detection, this may cause significant drops in detections, because of the high imbalance of benign and malware traffic. Solution may be lowering the forest decision threshold instead of requiring majority of votes to be positive. % no citacion found, however even scikit implements ROC curve this way. I think no citacion is fine here
This may increase the number of detections of the minority class, but often at the cost of precision of detections. Note that the thresholding is still difficult to apply to multi-class classification problems since different threshold has to be inferred and applied for each class separately.

The idea behind {\it Bayesian Tree Aggregation} (BTA) algorithm, pursued in this paper, is to make use of random sampling of the training set (i.e. bagging), sampled for each tree in the forest individually whenever a decision tree is trained. The aggregation makes use of prediction errors related to the performance on the unseen part of the training data (out-of-bag datasets). The additional information is used along with the Bayes theorem to handle imbalances and multi-class problems naturally. While the aggregation function is known \cite{kittler1998combining}, it is not wildly used and it was  generally not recognized to be well suited for imbalanced data. We show how it can be easily coupled with the training of random forests and what is the relation to the imbalanced class problem.
%mention that the aggregation function is generally not used because undesirable properties such as need for hold out dataset in the training set. Also possibly that the function is almost never used and we noticed the use case on imbalanced data

Even if many versions of the random forest algorithm were developed over the years, with the ambition to further improve random forest's general or domain-specific predictive performance \cite{schulter2013alternating,kontschieder2015deep,rainforth2015canonical,blaser2015random}, production quality code for most of these algorithms is not publicly available. Therefore, the original Random Forests \cite{Breiman:2001:RF:570181.570182} along with the Extremely Randomized Trees \cite{Geurts:2006:ERT:1132034.1132040} are still the most frequently used versions of the algorithm implemented in standard machine learning libraries \cite{scikit-learn,Meng:2016:MML:2946645.2946679}. An attractive property of BTA is that it may be implemented on top of models trained with these libraries.
%The algorithm does not alter the shape of individual trees, nor logic inside their nodes. During the training phase it only requires to perform an additional pass over the training data after the trees are inducted to compute the statistics from the out-of-bag datasets. 
This is particularly important in case of big data, since optimized state-of-the-art implementations may still be used to train the classifiers in the ensemble.

%Although the algorithm behind Bayesian Tree Aggregation is best suited for decision forests, because they already perform bagging as part of their training phase, it is not required that the weak classifiers in the ensemble are decision trees. In other words, it is entirely possible to use the algorithm together with ensembles of bagged classifiers of any type.

% advantages

\section{STANDARD MAJORITY VOTING}

Consider a classification problem with input space $\mathcal{X} = \mathbb{R}^M$ and output label space $\mathcal{Y} = \{1, \ldots, K\}$, and a decision forest with $T$ trees $\mathcal{T}_t : \mathcal{X} \mapsto \mathcal{Y}$. Each decision tree predicts a class label $\Tx{t} \in \mathcal{Y}$ for a given test sample $\vect{x} \in \mathcal{X}$ where $t \in \{1, \ldots, T\}$. In this notation, the predicted class label $y^*$, predicted by the forest can be obtained according to:

\begin{equation}
y^* = \argmax_{y \in \mathcal{Y}}\sum_{t = 1}^{T}I(\Tx{t} = y),
\end{equation}

where $I(\cdot)$ is the indicator function which returns $1$ if the condition in the argument is true, $0$ otherwise. \\

\section{BAYESIAN TREE AGGREGATION}

%Bayesian Tree Aggregation is very well suited for multi-class classification and datasets with imbalanced classes, because it leverages the information about class imbalance which the commonly used algorithms such as majority voting neglect.

The Bayesian Tree Aggregation (BTA) algorithm is inspired by the naive Bayes classifier. It computes the conditional probability of the label, $y$, of sample $\vect{x}$ given the predictions of individual trees. The final classification label $y^*$ of sample $\vect{x}$ is given by the maximal conditional probability, which can be expressed as:

\begin{equation}
\label{eq:max}
y^* = \argmax_{y \in \mathcal{Y}}P(y | \Tx{1}, \Tx{2}, \ldots, \Tx{T}).
\end{equation}

In accordance to the Bayes' theorem, we can rewrite~\eqref{eq:max} to the form:

\begin{equation}
y^* = \argmax_{y \in \mathcal{Y}}\frac{P(y) \cdot P(\Tx{1}, \Tx{2}, \ldots, \Tx{T} | y)}{P(\Tx{1}, \Tx{2}, \ldots, \Tx{T})}.
\end{equation}

Since the denominator does not depend on $y$, it can be ignored yielding:

\begin{equation}
y^* = \argmax_{y \in \mathcal{Y}}P(y) \cdot P(\Tx{1}, \Tx{2}, \ldots, \Tx{T} | y).
\end{equation}

To make the computation tractable, a common assumption is made that the  predictions of individual decision trees are independent given the true class label $y$. This approximation is in accordance with the aim of the decision-forest training algorithm, where randomness is introduced to reduce the correlation between individual trees. As a consequence, independence of errors between the trees can be assumed. This results in:

\begin{equation}
\label{eq:prod}
y^* = \argmax_{y \in \mathcal{Y}}P(y) \cdot \prod_{t = 1}^{T}P(\Tx{t} | y).
\end{equation}

The class $y^*$ is computed as the joint probability of individual tree predictions. That is, while the standard voting schemes directly aggregate predictions $\Tx{t}$ to make the final decision, in this case, in addition, also the probability $P(\Tx{t} | y)$ of a prediction for a given class is assessed.

As with the naive Bayes classifier, it is generally useful to transform the product to sum of logarithms to avoid problems with underflows in floating point arithmetics:

\begin{equation}
\label{eq:logFinal}
y^* = \argmax_{y \in \mathcal{Y}}\log P(y) + \sum_{t = 1}^{T}\log P(\Tx{t} | y).
\end{equation}

It is worth emphasizing that assumption of general independence between tree predictions $\Tx{i}$ and $\Tx{j}$ is not valid. Knowing the value of the prediction $\Tx{i}$ certainly provides information leading to a better estimate of $P(\Tx{j})$. Therefore, what is assumed is the \textit{conditional independence between tree predictions} given the true class label of sample $\vect{x}$, i.e. independence of errors.

\subsection{Estimation of underlying probabilities}
\label{sec:bta_estimation}

In order to decrease correlations between individual trees, each tree is trained only on a subset sampled from the training dataset (bagging) \cite{Breiman:2001:RF:570181.570182}. We make use of the part of the training dataset not seen during the training of a tree, i.e. the out-of-bag (OOB) dataset. Each tree is associated with it's own OOB dataset. In the standard setup, the size of the OOB dataset is approx. 37\% of the original training dataset size. Breiman \cite{brei94} has shown that if the training dataset and test dataset originate from the same underlying distribution then the OOB error on the training dataset is an accurate estimate of the generalization error.

% OOB error je lepsi estimate 

Conditional probability $P(\Tx{t} | y)$ for tree $t$ can be estimated from the confusion matrix computed from the out-of-bag dataset, not seen in the training phase of the tree $t$. Let $c_{k,l} \in \mathbb{N}$ be the element of the confusion matrix for tree $t$, where $k \in \mathcal{K}$ refers to the true class label and $l \in \mathcal{K}$ to the predicted class label. Then, $c_{k,l}$ equals the number of objects in the out-of-bag dataset that are classified as class $l$ but their true class label is $k$. The conditional probability can be computed as:
\begin{equation}
\label{eq:confmat}
P(\Tx{t} | y) = \frac{c_{y, \Tx{t}}}{N_y} = \frac{c_{y, \Tx{t}}}{\sum_{i = 1}^{K}c_{y, i}}.
\end{equation}
Probability $P(y)$ can be estimated from the prevalence of class $y$ in the training dataset. 

\subsection{Conditional probabilities smoothing}

Because the probabilities computed from the confusion matrices are merely estimates of the true underlying probabilities, it is reasonable to set the lower bound for the estimate to a small non-zero $\epsilon$ value rather than zero. Otherwise, the product in \eqref{eq:prod} would be zero if any of the posteriors \eqref{eq:confmat} were zero. A single decision tree would therefore be able to disable prediction of the whole forest for a given label. The value of $\epsilon$ is related to the confidence of the estimate and therefore it's value can generally be smaller with bigger training datasets. 

Kuncheva \cite{kuncheva2004combining} presents another way inspired by \cite{titterington1981comparison} of dealing with zeros in naive Bayes aggregation. Instead of using non-zero $\epsilon$ they use the formula:

\begin{equation}
\label{eq:confmat_kuncheva}
P(\Tx{t} | y) = \left(\frac{c_{y, \Tx{t}} + \frac{1}{c}}{N_y + 1}\right)^B,
\end{equation}

where $c$ is the number of classes and $B$ is a hyperparameter for which they suggest values: $0.5$, $0.8$ or $1$. Experiments in Section \ref{sec:security} show that the results are comparable with the $\epsilon$ method.

\subsection{Analysis of the algorithm}

The conditional probabilities $P(\Tx{t} | y)$ are conditioned on the true class label $y$. This causes their values to be independent of the class imbalance ratios because they are computed only from samples belonging to class $y$. This can be seen directly from \eqref{eq:confmat}, where the number of predictions of a class is normalized by the number of samples, $N_y$, in that class. Therefore, smaller classes require smaller number of predictions to achieve larger scores.

The information on the class imbalance is present in \eqref{eq:prod} in form of the prior $P(y)$. The importance of the prior decreases as the number of trees in the forest increases and the more trees agree on the same predicted class. % Comment why this is good?

%On the other hand, majority voting aggregates the votes from the individual trees in a non-probabilistic manner. Each tree's prediction is related to the probability $P(y | \vect{x})$ which is related to prior probability $P(y)$ by formula:
%
%$$
%P(y | \vect{x}) = P(y)\cdot\frac{P(\vect{x} | y)}{P(\vect{x})}
%$$
%
%Summing the predictions will therefore bias the result in favor of the majority class leading to higher precisions on minority classes but lower recalls.

% Put here an example with each tree 60 % certain of majority class, minority prevalence 1 %, and 100 trees? Maybe too long?

Additionally, because BTA does not work with votes from individual decision trees, but instead relies on probabilities computed on the out-of-bag (OOB) dataset, it is able to correct situations when samples from the OOB set belonging to some class $y_1$ are being consistently misclassified by a tree as class $y_2$. In this case, in the BTA estimation process the estimate of the posterior \eqref{eq:confmat} related to $y_1$ will get increased while for $y_2$ it will decrease.

% BTA aggregates the individual probabilities that originate from the evaluation of decision trees on out-of-bag datasets. This

\section{RELATED WORK}

Methods addressing the imbalanced class problem that can be used with decision forests focus mainly on preprocessing of training datasets by oversampling, undersampling or creating synthetic objects for the underrepresented class (e.g. SMOTE \cite{SMOTE}). Another common approach is cost-sensitive learning, which incorporates class weights to the learning process \cite{chen2004using}, but determining the correct costs is often difficult, especially in cases of extremely imbalanced data. Various reviews and alternatives
of these methods were published \cite{Krawczyk2016,galar2012review,lopez2013insight,chen2004using,he2009learning}. All of these methods can be used in combination with Bayesian tree aggregation, because they constitute different stages in the training algorithm.

In \cite{zhou2012loan,chen2017parallel} the focus is also put on the aggregation algorithm of trees in the random forest. The OOB is used to estimate tree weights in the voting process. Each tree weight represents error of a tree on the OOB dataset. The difference is that instead of simple reweighing of predictions of a tree, we build a probabilistic framework, where each prediction of a tree $t$ is associated with a vector of conditional probabilities $P(\Tx{t} | y)$ for classes $y \in {1, \ldots, K}$.

In Bayesian Model Averaging (BMA) \cite{hoeting1999bayesian} and related methods each model is weighted by the probability that the data were generated by the given model. However, it is difficult to use BMA with decision trees \cite{kim2012bayesian}. % why?

Kittler et al. \cite{kittler1998combining} include Bayesian decision aggregation in their work about classifier combining. However, they do not mention it in the context of bagged classifiers (such as Random forests) and therefore do not conveniently leverage the OOB datasets to estimate the necessary conditional probabilities. 

Kuncheva \cite{kuncheva2004combining} computes the confusion matrices from the dataset used to train the classifier, which is not a good estimate of the classifier's performance on unseen data. For example, due to unlimited depth of decision forests in experiments described in Section \ref{sec:security}, the number of misclassified objects in the training dataset is very close to zero. 

%Prior work concerned with analysis of HTTP logs has focused on unsupervised malware detection \cite{kohout2015unsupervised}, command-and-control server identification \cite{nelms2013execscent} and also on supervised learning with labels derived from domain blacklists \cite{franc2015learning,Bartos2015}. Franc et al. \cite{franc2015learning} focus on the problem of Multiple-Instance-Learning while Bartos et al. \cite{Bartos2015} focus on the domain adaption from the training to testing dataset considering that the distribution changes in time. %http://ecmlpkdd2017.ijs.si/papers/paperID193.pdf , https://publishup.uni-potsdam.de/opus4-ubp/frontdoor/deliver/index/docId/10094/file/httpsmalware.pdf

\section{DETECTION OF INFECTED USERS FROM NETWORK DATA}
\label{sec:security}
BTA was developed as a module for malware intrusion detection system with focus on network traffic, where the benign applications constitute majority of the network traffic. In this section, BTA will be compared against standard majority voting, available in standard machine learning libraries \cite{scikit-learn,Meng:2016:MML:2946645.2946679}, on a real-world dataset containing network proxy logs. Proxy logs record communication over HTTP(S) between single user and a single server. They consist of fields given by numerical values (e.g. transfered bytes), strings (e.g. URL, User-Agent, MIME type), categorical (HTTP status, ports) and other \cite{proxyLogs,Bartos2015}. The logs are bidirectional, therefore both directions of the communication are included in a single log. We are interested only in the non-encrypted communication (HTTP), therefore all proxy logs related to HTTPS communication were discarded from the experiments. Proxy logs were collected from more than 500 enterprise networks, ranging from small to large companies with tens of thousands of users.

Our NIDS is composed of two layers. First layer is an anomaly detection layer comprising more than 30 detectors, detecting anomalies according to empirical estimates of (conditional) probabilities such as P(country), P(domain$|$host), P(User-Agent$|$second level domain), time series analyses (models of user activity over time, detection of sudden changes in activity, identification of periodical requests, etc.), and HTTP specific detectors~\cite{anomaly2016}. Only 10\% of the most anomalous traffic is then propagated to the following classification layer, in which specific labels are assigned to some of the anomalies.

\subsection{Dataset description}
The labeling was performed on the level of contacted domains (it would be unfeasible to label each log separately). Labels were collected from available blacklists, other malware feeds from Collective Intelligence Framework (CIF)~\cite{farnham2013tools} or they were created by a human analyst.
Majority of proxy logs still remained unlabeled. Rather than keeping them out of the evaluations,  they were assumed to be benign and constitute the negative class. Therefore, true precision of the system may be higher than reported, because some of the false positives, which are in fact related to malware, may not be labeled yet.

\begin{figure}
\centering
\caption{Sizes of positive classes in the training and testing datasets. Note that the Y axis is in log-scale. The lower of the bars is shown in the front. Yellow color represents situation in which both training and testing dataset sizes for a given class are equal. Note the high imbalances present already in the positive malware-related classes.}
\label{fig:network_classes}
\includegraphics[width=1.0\linewidth]{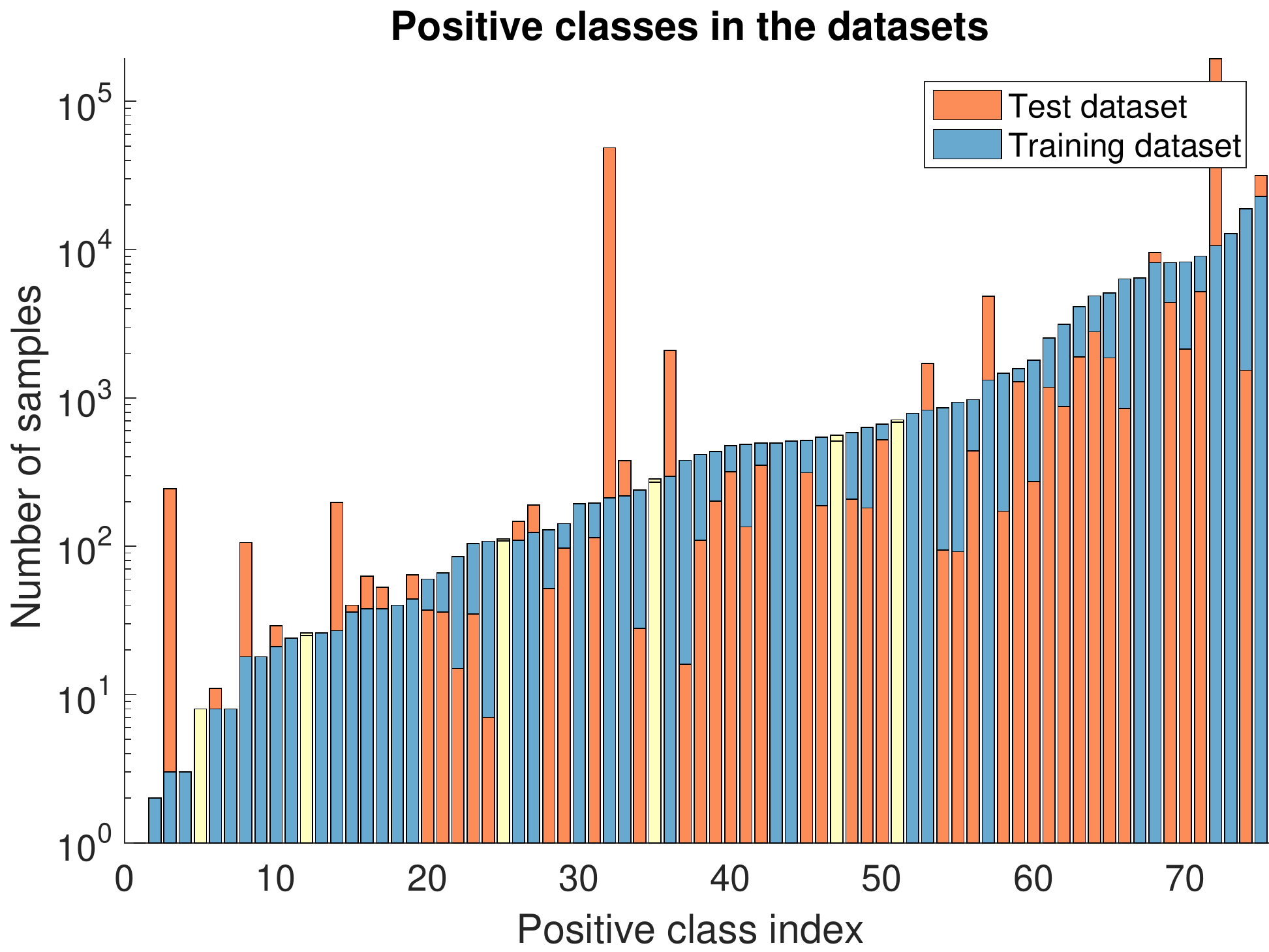}
\end{figure}

Malware was further categorized to classes. Each class is related to a traffic associated with different malware category, such as ransomware, trojan, click-fraud, exploit kit, exfiltration, ad-injector, etc. according to estimated risk level. Moreover, several subclasses for each category were defined based on the differences in the communication patterns of the malware. Our analysts were able to construct 75 different malware classes for 20 malware categories. The distribution of positive classes is shown in Figure~\ref{fig:network_classes}. Note the significant imbalance even between the positive classes alone.

% describe that it is worldwide traffic from many companies? Proprietary Cisco dataset?
Training dataset consists of proxy logs recorded during October and November 2015. Test data were collected in a single, busiest working day, Wednesday, 20th January 2016. Only top 10\% of samples, sorted according to the anomaly value are kept. Note that there is a time gap between both sets. This is necessary to properly address possible shift of the data in time.

From each proxy log, 357 features are extracted according to Bartos et al. \cite{Bartos2015}. The features are related to statistical properties of strings present in the URL, referer, User-Agent and information in the other proxy log fields such as: client/server port numbers, connection duration, number of bytes uploaded/downloaded, etc. Aim is to discover and learn communication patterns specific for malware.%The features are extracted from URL and referer and they are based on patterns that were shown to be able to identify malicious behavior such as: number of vovels in URL, number of special characters in URL, length of the longest stream of vowels, etc. Another set of features is extracted from general information that are available in each flow such as: client/server port numbers, HTTP status, connection length, number of bytes uploaded, User-Agent, etc.

For training, feature vectors related to benign traffic were uniformly downsampled, but all the malware samples were used. Overall, number of training instances was 3.5M, out of which 150k were attributed to malware. In the test set, no downsampling was performed yielding 10.9M feature vectors out of which 322k are related to malware. 

% cite my thesis?

% From thesis
%\begin{table}[!htb]
%\centering
%\caption{Example of features extracted from proxy log~\cite{Bartos2015}. The features in the right column are extracted from each part (protocol, second-level domain, tld, path, filename, query, fragment) of URL and referer.}
%\label{tab:network_features}
%\begin{tabular}{@{}l|l@{}}
%\arrayrulecolor{black}
%\hline
%\textbf{Features} & \textbf{Features on all URL parts + referer} \\ 
%\hline
%communication duration & total number of characters \\
%HTTP status & digit ratio \\
%User-Agent length & lower case ratio \\
%number of query parameters & upper case ratio \\
%number of bytes uploaded & vowel changes ratio \\
%number of bytes downloaded & has repetition of ’\&’ and ’=’ \\
%is URL in ASCII & starts with number \\
%client port number & number of non-base64 characters \\
%server port number & has a special character \\
%is URL encrypted & max length of consonant stream \\
%MIME-Type length & max length of vowel stream \\
%number of ’/’ in path & max length of lower case stream \\
%number of ’/’ in query & max length of upper case stream \\
%number of ’/’ in referer & max length of digit stream \\
%is second-level domain raw IP & ratio of a character with max occurrence \\
%\hline
%\end{tabular}
%\end{table}

\subsection{Evaluation metrics}
\label{sec:network_eval}

Even though the model classifies individual proxy logs, the evaluation is performed on the level of users. We assume that a user is infected if any of proxy logs related to his communication was labeled as malicious. Therefore, precision of a given class does say how many users, not proxy logs, that were identified as infected are truly infected.
Evaluation on the user level is necessary, because it truly reflects the perceived efficacy of the system by the customer, whose interest is mainly in the number of user devices that have to be maintained and not in specific network communication of a device.

The reported evaluation metrics are {\it precision} ($\frac{TP}{TP+FP}$), {\it recall} ($\frac{TP}{TP+FN}$) and {\it F-score} ($2 \cdot \frac{precision \cdot recall}{precision+recall}$). These metrics are the natural choice in the field when handling imbalanced datasets \cite{kotsiantis2006handling,saito2015precision}. 

Precision states how many positive detections that are shown at the output of the system to a human analyst are truly positive. From the end-user perspective it is crucial to have very high precision because investigation of individual detections requires significant portion of the analysts time. Notice the difference from {\it false positive rate} ($\frac{FP}{TN+FP}$) which is also commonly used. In cases of high class imbalance ratios, false positive rates tend to be very small and non-informative. In case of precision, the class imbalances are taken into account. Recall gives the fraction of users truly infected by malware and correctly detected by the classifier. Usually there exists a trade-off between precision and recall. F-score is the harmonic mean of precision and recall and it is commonly used to combine them into a single number.

The metrics are defined on binary classification problems and are usually reported only for the minority class, because performance on the majority class is often uninformative. This is extended to multi-class problem by computing the metrics for each minority class using the one-vs-all strategy. For conciseness, averages over the malware (minority) classes are reported. This approach is known as {\it macro-averaging} \cite{sokolova2009systematic}.

Since the precision is affected by the imbalance ratio, the negative class in the testing dataset can not be subsampled and has to contain all the proxy logs, which would the classifier see in the production environment after it was deployed.

\subsection{Hyperparameter setup}
\label{sec:network_hyper}
The underlying decision forests are trained using the standard algorithm proposed in \cite{Breiman:2001:RF:570181.570182} with hyper-parameters set to common values. The number of trees is set to $20$. Minimum number of samples required for further splitting set to $2$. The number of considered features in each node is set to $\sqrt{F}$, where $F$ is the overall number of features. Standard bagging is used. 

Section \ref{sec:bta_estimation} introduced two different algorithms for handling zero probabilities computed from the confusion matrices inside BTA. Both algorithms are tested in the experiment. The hyperparameter $\epsilon$ is set to $10^{-5}$ and the algorithm described in \eqref{eq:confmat_kuncheva} is tested with values $0.5$, $0.8$, and $1$ for the hyperparameter $B$, which are the values suggested in \cite{kuncheva2004combining}.

\subsection{Results}
Results of experiments for Majority Voting (MV) and Bayesian Tree Aggregation (BTA) are given in Table \ref{tab:network_dataset}. All the BTA variants outperform MV in terms of F-score, which combines precision and recall into a single number. While BTA $B = 0.5$ has slightly higher F-score than BTA $\epsilon$, the decrease in precision is lower for the latter while both offer a significant increase in recall in relation to majority voting.  

It is possible that some of the users identified as false positive can still be infected by malware, but we did not perform any further investigation of these errors because it would require significant investment of time from an expert human analyst. Therefore, the precision can be thought of as the lower bound of the true precision. Based on our requirements, 90\% precision is sufficiently high for a NIDS to be deployed in a production environment.

%The algorithms for handling zero probabilities computed from the confusion matrices within BTA, namely \eqref{eq:confmat_kuncheva} and replacement of zeros by $\epsilon$ perform very similarly without a significant difference in their performance. This non-significant difference between results indicates that the algorithm was not too sensitive to the values of these hyperparameters. 

\begin{table}
\caption{Results for the experiment on the network dataset. For BTA $\epsilon$, conditional probabilities that were equal to zero were replaced by the value of $\epsilon$. BTA $B$ is based on \eqref{eq:confmat_kuncheva}.}
\label{tab:network_dataset}
\centering
\begin{tabular}{@{}l|ccc@{}}
\hline
 & \thead{Precision} & \thead{Recall} & \thead{F-score} \\ \hline
\thead{BTA $\epsilon = 10^{-5}$} & $94.1 \%$ & $71.3 \%$ & $\thltb{81.1 \%}$\\
\thead{BTA $B = 0.5$} & $91.9 \%$ & $72.8 \%$ & $\thltb{81.2 \%}$\\
\thead{BTA $B = 0.8$} & $90.5 \%$ & $72.0 \%$ & $\thltb{80.2 \%}$\\
\thead{BTA $B = 1$} & $91.4 \%$ & $72.5 \%$ & $\thltb{80.9 \%}$\\
\thead{Majority Voting} & $96.9 \%$ & $64.4 \%$ & $\thltb{77.4 \%}$\\ \hline
\end{tabular}
\end{table}

\section{EXPERIMENTS ON PUBLIC IMBALANCED DATASETS}
\label{sec:evaluation}

In addition, we selected several popular and public datasets to evaluate BTA on a variety of imbalanced multi-class data, because the network dataset is proprietary and not publicly available. All of the datasets are available in a common {\it LibSVM} format at \cite{CC01a}. Except for dataset {\it dna}, the classes in the original datasets are balanced. To make the remaining datasets imbalanced, several classes were joined together to create a single majority class. The majority class was always created from the bottom {\it k} classes depending on their numeric label. Similar technique was used previously in \cite{galar2012review,chen2004using}. The properties of the datasets are summarized in Table \ref{tab:public_datasets}.

\begin{table}[th!]
\caption{Properties of the datasets used in evaluation. The column \textbf{\#Classes} contains the numbers of classes in the original datasets and the numbers of classes in the datasets modified to be imbalanced that were used in the evaluation. The Column {\it Majority prior} shows the prevalence of majority class in each dataset.}
\label{tab:public_datasets}
\centering
\setlength\tabcolsep{4pt}
\begin{tabular}{@{}r|ccccc@{}}
\hline
 & \thead{\#Train} & \thead{\#Test} & \thead{\#Classes} & \thead{Majority prior} & \thead{\# Features} \\ \hline
\thead{usps} & 7291  & 2007 & 10 / 3 & 84 \% & 256 \\
\thead{dna} & 1400 & 1186 & 3 / 3 & 53 \% & 180 \\
\thead{letter} & 15000 & 5000 & 26 / 7 & 77 \% & 16 \\
\thead{satimage} & 3104  & 2000 & 6 / 3 & 73 \% & 36 \\
\thead{aloi} & 98000  & 10000 & 1000 / 200 & 80 \% & 128 \\
\thead{mnist} & 60000 & 10000 & 10 / 3 & 80 \% & 780 \\ \hline
\end{tabular}
\end{table}

\subsection{Evaluation metrics}

The same metrics as in Section \ref{sec:network_eval} are used. The experiment is designed in a similar way to the experiment on network data. The majority class is treated in the same way as the negative class and the minority classes are treated as the positive classes. Metrics are again computed in the one-vs-all manner and the reported numbers are the averages calculated using the macro-averaging strategy.

\subsection{Hyperparameters setup}

The hyperparameters were set in the same way as in the experiments on network data described in Section \ref{sec:network_hyper} with exceptions that the number of trees was set to $100$ because the datasets are smaller which allows faster training and it is a common choice.

Zero probabilities computed from the confusion matrices inside BTA were replaced by $\epsilon = 10^{-5}$ in all experiments.

\subsection{Results}
Results are given in Table \ref{tbl:results}. Majority voting has better precision and Bayesian tree aggregation has better recall on all datasets. Since we are interested mainly in the trade off between precision and recall, main focus is placed on the F-score. In terms of F-score, BTA has the best performance on all datasets except for the {\it usps} dataset, where the performance is comparable to that of majority voting. Note that every experiment was repeated $10$ times because the training of decision forests contains randomness. The averages and standard deviations for each metric are reported.

% elaborate a bit more?

\begin{table*}[ht!]
\caption{Bayesian tree aggregation compared with majority voting. Each experiment was repeated ten times and the reported values are means and standard deviations of the results.}
\label{tbl:results}
\centering
\begin{tabular}{r|cc|cc|cc}
\hline
         & \multicolumn{2}{c|}{\thead{Precision}} & \multicolumn{2}{c|}{\thead{Recall}} & \multicolumn{2}{c}{\thead{F-score}} \\
\thead{Dataset} & \thead{MV}         & \thead{BTA}        & \thead{MV}       & \thead{BTA}       & \thead{MV}        & \thead{BTA} \\ \hline
\thead{usps} & $\thlt{0.960 \pm 0.005}$ & $0.874 \pm 0.007$ & $0.849 \pm 0.006$ & $\thlt{0.920 \pm 0.005}$ & $\thltb{0.900 \pm 0.005}$ & $0.897 \pm 0.005$\\
\thead{dna} & $\thlt{0.934 \pm 0.006}$ & $0.920 \pm 0.005$ & $0.907 \pm 0.010$ & $\thlt{0.943 \pm 0.003}$ & $0.920 \pm 0.006$ & $\thltb{0.932 \pm 0.003}$\\
\thead{letter} & $\thlt{0.990 \pm 0.002}$ & $0.965 \pm 0.002$ & $0.914 \pm 0.002$ & $\thlt{0.963 \pm 0.002}$ & $0.950 \pm 0.002$ & $\thltb{0.964 \pm 0.002}$\\
\thead{satimage} & $\thlt{0.920 \pm 0.002}$ & $0.879 \pm 0.004$ & $0.836 \pm 0.004$ & $\thlt{0.898 \pm 0.006}$ & $0.876 \pm 0.002$ & $\thltb{0.889 \pm 0.003}$\\
\thead{aloi} & $\thlt{0.990 \pm 0.001}$ & $0.966 \pm 0.001$ & $0.851 \pm 0.003$ & $\thlt{0.962 \pm 0.002}$ & $0.908 \pm 0.002$ & $\thltb{0.961 \pm 0.001}$\\
\thead{mnist} & $\thlt{0.986 \pm 0.001}$ & $0.938 \pm 0.001$ & $0.889 \pm 0.004$ & $\thlt{0.947 \pm 0.002}$ & $0.935 \pm 0.002$ & $\thltb{0.943 \pm 0.001}$\\      
\hline
\end{tabular}
\end{table*}

\section{CONCLUSIONS}

Bayesian Tree Aggregation (BTA) can be used in place of any ensemble aggregation algorithm, but it is best-suited for imbalanced multi-class problems, where detection of the minority class is of high importance. BTA utilizes the out-of-bag dataset to estimate the prediction errors of individual decision trees. The method employs Bayesian reasoning to combine the individual votes. The method affects only the prediction of the ensemble and not the topology of individual trees. Therefore, it can be easily used with any state-of-the-art implementation for induction of decision forests. This property is particularly useful on distributed platforms such as Apache Spark, where highly optimized version of the algorithm already exist and implementing the method from scratch would be difficult.

The method was evaluated on the task of intrusion detection. Real-world dataset was collected. Additionally, the algorithm was also tested on several popular public datasets. In majority of the cases, the method was consistently able to achieve higher average F-scores for minority classes of interest.

% incorporate experiment without OOB into paper?

%While it might be difficult to obtain the out-of-bag datasets, that were used for training of the individual trees by the third-party library, during our experiments we have observed that it is possible to achieve same results with newly sampled bagged datasets in place of the out-of-bag datasets.

% \vfill\pagebreak

%\bibliography{mybase}{}
%\bibliographystyle{IEEEtran}

\bibliographystyle{IEEEbib}
\bibliography{mybase}{}

%TODO: research the correct bibliography style and also the citation style.

\end{document}